\pdfoutput=1

\documentclass[11pt]{article}

\usepackage[final]{mtsummit25}
\usepackage{copyright}

\usepackage{times}
\usepackage{latexsym}
\usepackage{enumitem} 

\usepackage[T1]{fontenc}
\usepackage{booktabs}
\usepackage[utf8]{inputenc}
\usepackage{adjustbox} 

\usepackage{CJKutf8}
\usepackage{makecell}
\usepackage{multirow}
\usepackage{microtype}

\usepackage{inconsolata}
\usepackage{amsmath} 
\usepackage{graphicx}

\title{Decoding Machine Translationese in English-Chinese News:

LLMs vs. NMTs}

\author{
 \textbf{Delu Kong\textsuperscript{1,2}}
 \textbf{ and Lieve Macken\textsuperscript{2}}
\\
\\
 \textsuperscript{1}School of Foreign Studies, Tongji University, Shanghai, 200092, China \\
 \textsuperscript{2}Language and Translation Technology Team, Ghent University, Ghent, 9000, Belgium
\\
 \small{
   \textbf{Correspondence:} \href{mailto:kongdelu2009@hotmail.com}{kongdelu2009@hotmail.com}
 }
}

\begin{document}

\begin{CJK}{UTF8}{gbsn} 

\maketitle
\begin{abstract}

This study explores Machine Translationese (MTese) — the linguistic peculiarities of machine translation outputs — focusing on the under-researched English-to-Chinese language pair in news texts. We construct a large dataset consisting of 4 sub-corpora and employ a comprehensive five-layer feature set. Then, a chi-square ranking algorithm is applied for feature selection in both classification and clustering tasks. Our findings confirm the presence of MTese in both Neural Machine Translation systems (NMTs) and Large Language Models (LLMs). Original Chinese texts are nearly perfectly distinguishable from both LLM and NMT outputs. Notable linguistic patterns in MT outputs are shorter sentence lengths and increased use of adversative conjunctions. Comparing LLMs and NMTs, we achieve approximately 70\% classification accuracy, with LLMs exhibiting greater lexical diversity and NMTs using more brackets. Additionally, translation-specific LLMs show lower lexical diversity but higher usage of causal conjunctions compared to generic LLMs. Lastly, we find no significant differences between LLMs developed by Chinese firms and their foreign counterparts.

\end{abstract}

\section{Introduction}

A recent, striking report -- with an arguably sensational title -- proclaims a groundbreaking milestone for LLMs in the field of machine translation (MT): \textit{``Machine Translation is Almost a Solved Problem''}\footnote{\url{https://www.economist.com/science-and-technology/2024/12/11/machine-translation-is-almost-a-solved-problem}}. Although the article's perspective is primarily forward-looking, with a clear acknowledgment on the enduring value of human translation, its message to the public, as the title suggests, is rather obvious: With the help of LLMs, MT currently appears to be nearing perfection. But is it? 

Multiple studies showed that LLMs have revolutionized the way we approach language translation, reaching to an unprecedented level of accuracy, contextual understanding, and fluency~\cite{jiao_is_2023,peng-etal-2023-towards,wang-etal-2023-document-level,enis_llm_2024}. It even outperformed some specialized NMT systems under a fine-grained evaluation setting~\cite{manakhimova-etal-2023-linguistically}. Further compelling evidence of the benefits of LLMs is reflected in the WMT24 finding summary~\cite{kocmi_findings_2024}, which clearly demonstrate their dominance in the competition. Most of the top-performing systems were LLM-based, with standout models, like Claude-3.5-sonnet, achieving leading positions across multiple language pairs. 

Despite their advantages, several limitations persist. Notably, LLMs often face challenges with low-resource languages~\cite{enis_llm_2024}, in explaining, practicing and translating sophisticated concepts~\cite{qian_exploring_2024}, and addressing gender bias issues associated with vocabulary options~\cite{stafanovics_mitigating_2020}. An area that remains largely underexplored is the influence of machine translationese.
While MTese has been demonstrated in NMT output~\cite{vanmassenhove-etal-2019-lost,vanmassenhove_machine_2021}, it has not yet been fully investigated in the context of LLMs. 

MTese can subtly influence the readability, naturalness, and even credibility of news articles, potentially shaping public perceptions. Studying MTese is critical from several perspectives. In education, MT has been widely utilized in second language acquisition, with MT output sometimes even regarded by students as ``expert''~\cite{rowe_google_2022}. However, MTese may potentially impact the authenticity of learning materials, raising concerns about its influence on learners' lack of exposure to genuine linguistic patterns. In language evolution, it provides insights into how machine-mediated communication might drive changes in linguistic norms, and whether MTese would also plays a part in the course, such as influencing language complexity~\cite{cristea_archaeology_2024}. In literary translation, MTese poses a potential hindrance to the creativity and linguistic richness of literary translation, continuously challenging the long-debated concept of ``human parity''~\cite{poibeau__2022} from a stylistic perspective.

Against this background, we have chosen MTese as the focus of this study, specifically exploring its manifestations and differences in NMTs and LLMs for the linguistically distant language pair of English to Chinese~(E2C).

\section{Related work}

First introduced by \citet{gellerstam_translationese_1986}, the concept of translationese describes the systematic influence of a source language on the target language during translation. When applied to MT, \citet{daems_translationese_2017} emphasize the pivotal role of MTese in shaping the characteristics of post-edited texts, analyzing 55 linguistic features ranging from POS tags to dependency parsing.

Expanding on this, \citet{toral_post-editing_2018} explore lexical density and diversity, revealing that post-edited literary MTs tend to be more simplified, normalized, and influenced by the source text compared to human translations~(HTs), where MT outputs exhibit lower lexical density than HTs, with the neural system showing even lower density than those from the statistical system. However, \citet{castilho_what_2019} report contrasting findings from a different genre. For news texts, MTs show slightly higher lexical density and richness than HTs, whereas for literary texts, MTs demonstrate slightly lower lexical density but comparable lexical richness to HTs. 

In a similar vein, \citet{loock_no_2020} investigate MTese by analyzing linguistic deviations in English-to-French MT texts compared to original, untranslated texts, providing a broader perspective on the systematic over-representation of linguistic features and their implications for translator training and post-editing practices. \citet{de_clercq_uncovering_2021}, working on the same language pair, used 22 linguistic features to distinguish between the original and MTed French. They showed that average sentence length and four features related to formulaicity could discriminate between original and MTed French. 

However, for linguistically distant language pairs like E2C, research on MTese remains relatively sparse. \citet{jiang_corpus-based_2022} examine a corpus of English translations of modern Chinese literary texts, including texts translated by NMT and humans. They confirm the presence of translationese in both human and machine translations compared to original texts in some coherence metrics. A recent study by \citet{niu_does_2024} revealed that simplification is a notable characteristic of NMT texts across genres in the E2C direction, such as a loss of lexical complexity. 

A general conclusion drawn from the works above is that translations produced by MT engines consistently exhibit a loss of lexical and syntactic richness~\cite{vanmassenhove-etal-2019-lost, castilho_post-editese_2022}. Studies tend to apply fine-grained linguistic features to reveal consistent distributional patterns. Similar phenomena are also observed across various language pairs.

Despite these findings, it remains unclear whether LLMs exhibit distinct features of MTese or whether their outputs can be reliably differentiated from those of NMT engines, particularly in general text types as news discourse. Therefore, our study addresses this gap by focusing on the distant language pair of E2C, emphasizing general news texts, and constructing a larger and more comprehensive dataset and feature set for analysis. We adopt the study design  of \citet{loock_no_2020} and \citet{de_clercq_uncovering_2021}, and compare MT~\footnote{In this paper, the term MT is used as a superordinate term for both NMT and LLM translation, while NMT and LLM could also be treated as distinct categories.} outputs generated by different systems with original texts. 

We address the following research questions:
\begin{itemize}[noitemsep, topsep=0pt]
    \item RQ1: Does MTese exist in E2C MTed news texts (NMTs and LLMs)? If so, which linguistic features contribute most to this distinction?
    \item RQ2: How do LLMs differ from NMTs in their manifestation of MTese across linguistic features?
    \item RQ3: Do translation-specific and generic LLMs differ from each other? Additionally, how do LLMs developed by Chinese companies compare with those developed by foreign companies in this regard? 
\end{itemize}

\section{Methodology}
\subsection{Dataset}

The dataset used in this study encompasses four corpora (detailed information is shown in Table \ref{tab: dataset}), representing original Chinese texts and E2C MTed news texts. The original Chinese news corpus is sourced from two authoritative outlets, \textit{People's Daily}\footnote{\url{http://www.people.com.cn/}} (人民日报) and \textit{Xinhua News}\footnote{\url{http://www.xinhuanet.com/}}(新华网), while the original English news corpus includes articles from reputable platforms like \textit{The Economist}\footnote{\url{http://www.economist.com}} and \textit{The Guardian}\footnote{\url{https://www.theguardian.com}}. The selected corpora consist exclusively of news texts published after 2022 to avoid the potential influence of outdated news that may have been incorporated into the training data of LLMs. The sample length within each original corpus is maintained around 900 words in average for both Chinese and English. 

All texts underwent careful preprocessing, including cleaning, denoising, part-of-speech (PoS) tagging, and dependency (Dep) tagging. Chinese is a language without explicit word boundaries, which requires word segmentation in advance. To achieve state-of-the-art performance, we utilized the Language Technology Platform (LTP)\footnote{\url{https://github.com/HIT-SCIR/ltp}}, a comprehensive natural language processing toolkit~\cite{che_n-ltp_2021}. We used LTP's advanced deep learning models (Base2) to perform word segmentation, PoS tagging, and syntactic analysis. Its reported performances reach 99.18\%, 98.69\%, and 90.19\% for segmentation, PoS tagging, and Dependency parsing, respectively.\footnote{\url{https://github.com/HIT-SCIR/ltp/blob/main/README.md}}

The MT data are generated by translating OEN articles into Chinese using each engine on a one-by-one basis, processing each text individually with each engine, thereby minimizing potential interference from varying text topics. As a result, each MT engine produces approximately 200 translations\footnote{It should be noted that several LLMs did not translate all 200 English news~(as in Table \ref{tab: dataset}). Some news articles remain untranslated due to ``unsafe content'' warnings, primarily involving topics related to war or politics. Even though we stated clearly in our prompt that they do not contain any unsafe content.~(See Appendix \ref{sec:appendix a})}. The dataset includes five NMT engines, comprising three international systems (Google Translate, DeepL, and Microsoft Translator) and two Chinese-developed systems (Baidu Translate and Youdao Translate). Additionally, six LLMs are incorporated, including models developed by Chinese firms (Kimi and ChatGLM), one tailored for MT-specific applications (TowerInstruct), and leading models such as ChatGPT, Claude, and Gemini. All these systems represent SOTA engines on the LLM arena\footnote{\url{https://lmarena.ai/}} at the time of the experiment (October to December 2024).

The MT process for LLMs involves prompt engineering, with prompt design following Andrew Ng’s course guidelines\footnote{ \url{https://learn.deeplearning.ai/courses/chatgpt-prompt-eng}} and the CRISPE framework\footnote{\url{https://github.com/mattnigh/ChatGPT3-Free-Prompt-List}}. This approach resulted in a standardized and structured user instruction, which was consistently applied across all engines during the translation process. The full prompt is provided in Appendix \ref{sec:appendix a} for reference.

To address the potential issue of the OCN corpus having insufficient coverage and variability when limited to the same sample size (200) as other sub-corpora, we adopted the corpus structure outlined in~\citet{de_clercq_uncovering_2021} and increased the size of the OCN by incorporating more original Chinese news to 2,000 texts in total. This expansion ensures a comparable dataset size between OCN and MTs. Also, we did not impose strict limitations on specific topics within the news genre. Constraining the dataset to a particular domain could lead to data scarcity, as certain topics may not be consistently available over the given period. To maintain balance and comparability, we ensured that all selected news articles were consistent in terms of lexical length and time period.

\begin{table*}[!h]
    \centering\small
    \begin{tabular}{l l l l l r r r}
    \toprule
    \textbf{Corpus} & \textbf{Type} & \textbf{Abbr.} & \textbf{Engine} & \textbf{Acquisition} & \textbf{Texts} & \textbf{Token} & \textbf{Type} \\
    \midrule
    \multirow{2}{*}{\textbf{Original}} & Orig. Chi. News & OCN & - & WebCrawl & 2,000 & 1,685,526 & 67,082 \\
     & Orig. Eng. News & OEN & - & WebCrawl & 200 & 190,572 & 20,459 \\
    \midrule
    \multirow{5}{*}{\textbf{NMTs}} & GoogleTranslate & NGT & - & API & 200 & 171,448 & 15,288 \\
     & DeepL & NDL & - & API & 200 & 177,866 & 15,104 \\
     & MicrosoftTranslator & NMS & - & API & 200 & 172,026 & 14,586 \\
     & BaiduTranslate$^{*}$ & NBD & - & API & 200 & 174,962 & 14,323 \\
     & YoudaoTranslate$^{*}$ & NYD & - & API & 200 & 174,504 & 16,515 \\
   
    \midrule
    \multirow{6}{*}{\textbf{LLMs}} & ChatGPT & LCG & GPT-4o & Web & 200 & 159,015 & 14,947 \\
     & Claude & LCL & 3.5-sonnet & Web & 200 & 170,236 & 15,123 \\
     & Gemini & LGM & 1.5-flash & API & 189 & 166,631 & 14,777 \\
     & Kimi$^{*}$ & LKM & moonshot & API & 185 & 145,976 & 13,225 \\
     & ChatGLM$^{*}$ & LGL & GLM-4-plus & API & 178 & 145,700 & 14,907 \\ 
     & TowerInstruct\textsuperscript{†} & LTO & 7B-v0.2 & OpenSource & 200 & 175,681 & 15,398 \\
     \midrule
     \textbf{MTs} & In total & MTs & NMTs + LLMs & - & 2,152 & 1,834,045 & 31,863 \\
    \bottomrule
    \end{tabular}
\caption{Overview of the datasets used in this study. Engines marked with an asterisk (*) are primarily trained and tested in mainland China, while the engine in the LLMs marked with a dagger (†) represents an translation-specific model.}
\label{tab: dataset}
\end{table*}

\subsection{Feature set}

This section outlines the feature set used in this study. The primary aim of this study is to quantitatively compare different linguistic features across original and MT texts. 

Based on the principles of constructing feature sets for translationese studies\cite{volansky_features_2013} and referring to previous research~\cite[See][]{huang_wei_application_2009, lynch_translators_2018, toral-2019-post,de_clercq_uncovering_2021}, the following section presents the feature set used in this study. A brief feature summary can be viewed in Table~\ref{tab:features} (in Appendix \ref{sec:appendix b}). All together, we have employed 236 features in this study. It should be noted that all features are represented as ratios or weighted measures to mitigate the influence of sample size differences and ensure comparability across texts.

\paragraph{Lexical features}
General lexical features involves common lexical characteristics such as Type-Token Ratio (TTR). The purpose of these features is to provide an overview of lexical usage in terms of diversity, complexity, and richness. PosTag-based features are derived from the annotation tag set of the LTP platform\footnote{\url{https://ltp.ai/docs/appendix.html\#id2}}. For instance, the proportions of nouns or verbs. 

\paragraph{Syntactical features}
General syntactical features focus on capturing broad syntactic patterns and sentence structures in the text, such as average words per sentence. DepTag-based features are built upon the dependency tag set of the LTP platform\footnote{\url{https://ltp.ai/docs/appendix.html\#id5}}, which identifies the dependency role of each word in a sentence, such as the ratio of verb-object (VOB) and attributive modifiers (ATT). 

\paragraph{Readability features}
Nine readability features proposed by \citet{lei_alphareadabilitychinese_2024}\footnote{\url{https://github.com/leileibama/AlphaReadabilityChinese}} are included, evaluating lexical, syntactic, and semantic variability to assess the text's difficulty and comprehensibility for the target audience. Complementing these are 4 concreteness features measuring lexical concreteness based on \citet{xu_concretenessabstractness_2020}. 

\paragraph{Translatibility features}

The translatability features evaluate linguistic coherence  and translation quality between English and Chinese texts through five features: completeness, foreignness, code-switching, abbreviation, and untranslated. Core features such as completeness check for untranslated English sentences longer than three words. Foreignness calculates the ratio of English to Chinese characters. 

\paragraph{N-POS-gram features}
To ensure that the feature set remained content-independent and focused on grammatical patterns rather than topical content, we employed N-PoS-grams (with N ranging from 1 to 3) instead of lexical n-grams. These features capture sequences of part-of-speech tags to highlight grammatical collocations across texts. To refine the selection and reduce the influence of highly frequent but less informative elements (e.g., function words), we used the The Lancaster Corpus of Mandarin Chinese (LCMC\footnote{\url{https://www.lancaster.ac.uk/fass/projects/corpus/LCMC/}}) as a reference corpus for comparison. For consistency, the LCMC corpus was re-tagged using the same LTP tools to ensure an aligned PoS tag set.

\subsection{Algorithms}

\subsubsection{Feature selection}

To reduce complexity, minimize feature noise, and improve experimental efficiency, a chi-square (\(\chi^2\)) ranking-based feature selection method is employed in both classification and clustering experiments. Features are ranked based on \(\chi^2\) values, and the top-\(k\) features are selected, where \(k = 30\). If the total number of features in a specific category is less than 30, all available features are retained. This feature selection process mainly reduces the lexical features, N-POS-gram features, and the combined ``all features'' set in classifying and clustering. 

\subsubsection{Classification experiment}

The classification experiment is structured based on the hierarchical levels of feature sets. First, classification is conducted using individual feature level, followed by classification using all feature levels. The experiments are organized into the following comparison groups: (1) OCN vs.~MTs, where MTs include both NMTs and LLMs subgroups; (2) OCN vs.~NMTs and OCN vs. LLMs; (3) LLMs vs.~NMTs; and (4) intra-group classifications within the NMTs and LLMs.  

Five classifiers, Naïve Bayes, Logistic Regression, Support Vector Machine (SVM), Decision Tree, and Random Forest, are employed, and the average classification performance is calculated across these classifiers to provide a balanced result. SVM uses a linear kernel, while the other classifiers follow default settings. Referring to~\citet{rahman_commentclass_2024}, the performance of the ensemble classifier is evaluated using Accuracy (ACC) and F1 scores, computed as follows:
\[
\text{ACC}_{\text{avg}} = \frac{1}{N} \sum_{i=1}^{N} \text{ACC}_i, \quad 
F1_{\text{avg}} = \frac{1}{N} \sum_{i=1}^{N} F1_i
\]

Where \( \text{ACC}_i \) and \( F1_i \) are the Accuracy and F1 scores for the \(i\)-th classifier, and \(N\) is the total number of classifiers. All classification tasks, except intra-group classifications within the NMTs and LLMs groups, are binary classification tasks.

\subsubsection{Clustering experiment}

The clustering experiment employs the \(k\)-means algorithm to cluster the data into three categories: OCN, LLMs, and NMTs. The number of clusters (\(k\)) is set to 3, and the Euclidean distance is used to measure the similarity between data points. The top-k significant features selected in prior analysis, are utilized as the feature set for clustering. 

To evaluate the clustering performance, the Adjusted Rand Index (ARI) is used as the primary metric. ARI measures the similarity between the clustering results and the ground truth labels, adjusted for chance, providing an objective assessment of clustering quality~\cite{warrens_understanding_2022}. Additionally, Python's Plotly library is employed to generate interactive clustering plots. This approach complements the classification methods, offering an intuitive visualization of the relationships between categories.

\section{Results}
\subsection{Classification}

\begin{table*}[h]
\centering\small
\renewcommand{\arraystretch}{1.2} 
\setlength{\tabcolsep}{2pt} 
\begin{tabular}{cccccccc}
\toprule
\textbf{\makecell{Feature\\Level}} & \textbf{Metrics} & \textbf{OCN-MTs} & \textbf{OCN-NMTs} & \textbf{OCN-LLMs} & \textbf{LLMs-NMTs} & \textbf{\makecell{NMTs\\(Intra-group)}} & \textbf{\makecell{LLMs\\(Intra-group)}} \\
\midrule
\multirow{3}{*}{Lexical} & ACC & 97.66\% & 97.77\% & 97.33\% & 61.91\% & 30.04\% & 30.26\% \\
 & F1 & 0.9766 & 0.9751 & 0.9714 & 0.6144 & 0.2893 & 0.2966 \\
 & C/T & 4054/4152 & 2932/3000 & 3067/3152 & 1332/2152 & 300/1000 & 348/1152 \\
\midrule
\multirow{3}{*}{Syntactical} & ACC & 98.46\% & 98.23\% & 98.23\% & 60.53\% & 36.72\% & 31.89\% \\
 & F1 & 0.9846 & 0.9801 & 0.9809 & 0.5736 & 0.3574 & 0.3022 \\
 & C/T & 4087/4152 & 2946/3000 & 3096/3152 & 1302/2152 & 367/1000 & 367/1152 \\
\midrule
\multirow{3}{*}{Readibility} & ACC & 86.87\% & 87.61\% & 85.41\% & 55.31\% & 19.44\% & 25.49\% \\
 & F1 & 0.8683 & 0.8603 & 0.8426 & 0.5452 & 0.1896 & 0.2437 \\
 & C/T & 3607/4152 & 2628/3000 & 2691/3152 & 1190/2152 & 194/1000 & 293/1152 \\
\midrule
\multirow{3}{*}{Translatibility} & ACC & 70.66\% & 78.73\% & 78.32\% & 58.26\% & 26.36\% & 20.73\% \\
 & F1 & 0.6265 & 0.6543 & 0.6618 & 0.4985 & 0.2146 & 0.1515 \\
 & C/T & 2933/4152 & 2362/3000 & 2468/3152 & 1253/2152 & 263/1000 & 238/1152 \\
\midrule
\multirow{3}{*}{N-POS-gram} & ACC & 97.27\% & 96.51\% & 96.73\% & 62.49\% & 24.26\% & 24.32\% \\
 & F1 & 0.9603 & 0.9469 & 0.9500 & 0.5287 & 0.2031 & 0.1301 \\
 & C/T & 3987/4152 & 2865/3000 & 3008/3152 & 1322/2152 & 232/1000 & 212/1152 \\
\midrule
\multirow{3}{*}{All Features} & ACC & 98.92\% & 98.84\% & 98.69\% & 69.38\% & 42.10\% & 35.59\% \\
 & F1 & 0.9891 & 0.9870 & 0.9859 & 0.6902 & 0.4136 & 0.3475 \\
 & C/T & 4106/4152 & 2965/3000 & 3110/3152 & 1492/2152 & 421/1000 & 409/1152 \\

\bottomrule
\end{tabular}
\caption{Performance metrics across feature levels and groups. ACC refers to Accuracy, F1 is a balanced score of precision and recall, while C/T stands for Correctly classified sample / Total samples. }
\label{tab: class}
\end{table*}

Table~\ref{tab: class} presents the results across feature levels and groups. We observe two tendencies: 

(1) OCN versus other groups consistently achieves the highest accuracy (around 99\% in all feature categories), regardless of whether MT is combined into one group or separated into NMTs and LLMs. Then the performance declines for LLMs-NMTs comparisons (around 70\% ACC). The lowest accuracy is observed in intra-group comparisons for both LLMs and NMTs, dropping to below 50\%, lower than random distribution baseline. 

(2) As for feature categories, lexical, syntactical and N-POS-gram features make the most significant contributions to the classification performance. Take OCN-MTs as an example, the three sets of features all reach to more than 97\%. In contrast, readability and translatability features show limited contributions, with accuracies of 86.87\% and 70.66\% respectively. Combined all features yields the highest overall accuracy (98.92\%), which demonstrates the complementary effects of integrating multiple feature categories. 

\begin{figure}[h]
    \centering
    \includegraphics[width=1\linewidth]{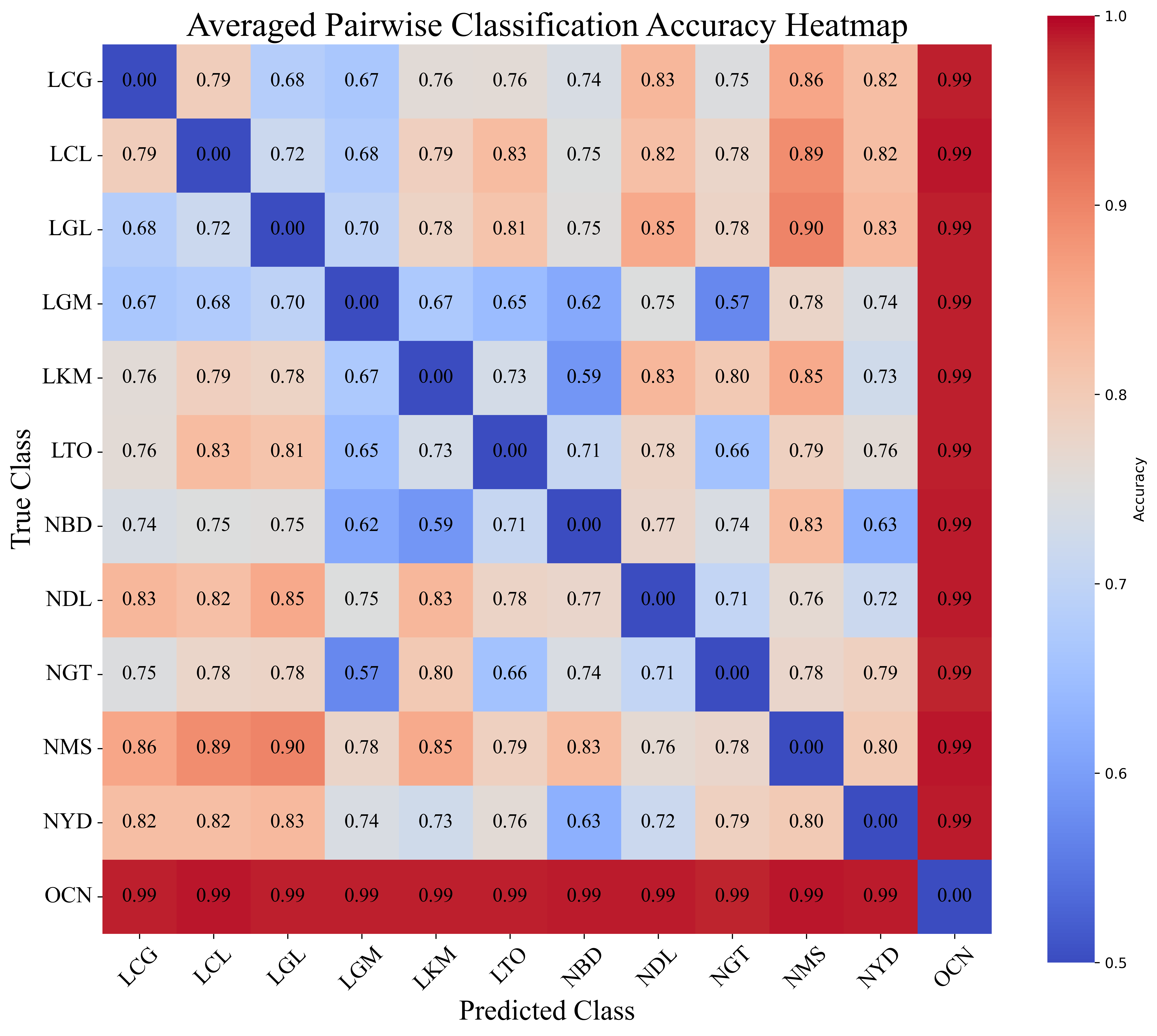}
    \caption{Pair-wise comparison of different MT engines based on 5 averaged classifiers and top 30 salient features}
    \label{fig:heatmap}
\end{figure}

Figure~\ref{fig:heatmap} presents a pairwise classification accuracy heatmap to provide a visualized plot and a fine-grained classifying result, using the top-30 salient features from all feature levels. The classification performance is averaged across the five above-mentioned classifiers. For each pair of classes, ACC values are computed and aggregated to produce the final heatmap.
It reveals three main results:

(1) The deep red along the OCN comparisons highlights its distinctiveness, achieving near-perfect classification accuracy (avg. ACC is close to 0.98) against both LLMs and NMTs. 

(2) A similar trend is found in both LLMs and NMTs intra-groups, reflected in the predominantly blue and light orange colours, which stand for 0.6 - 0.8 ACC according to the heatmap legend. For LLMs, ACC ranges from 0.65 to 0.83 (avg. 0.73). Similarly, NMTs exhibit ACC ranging from 0.63 to 0.83~(avg. 0.75)\footnote{Compared with Table.~\ref{tab: class}, the higher scores in pairwise models are due to binary classification tasks, which reduces task complexity and better captures discriminative features, whereas multi-class task involves increased feature overlap and requires generalization across all categories.}.

(3) ACC between LLMs and NMTs is slightly higher (avg. 0.77), indicating more distinct differences between these two groups. The lowest ACC is found between NGT and LGM (avg. 0.57), perhaps due to similar training data since they are both developed by Google\footnote{There exist a possibility of incorporating LLM technology into commercial NMTs, but the specific technical details remain unknown when the company does not disclose further information. So we only select NMT engines as ``pure'' as possible in our study. For example, the API we use for NGT is v2 (See \url{https://cloud.google.com/translate/docs/editions}), which is an NMT engine, rather than v3, which includes LLM.}. And the highest is found between NMS and LGL (0.90). Notably, NMS stands out with a slightly higher classification ACC against other LLMs (around 0.84). 

\subsection{Clustering}

\begin{figure*}[h]
    \centering
    \includegraphics[width=1\linewidth]{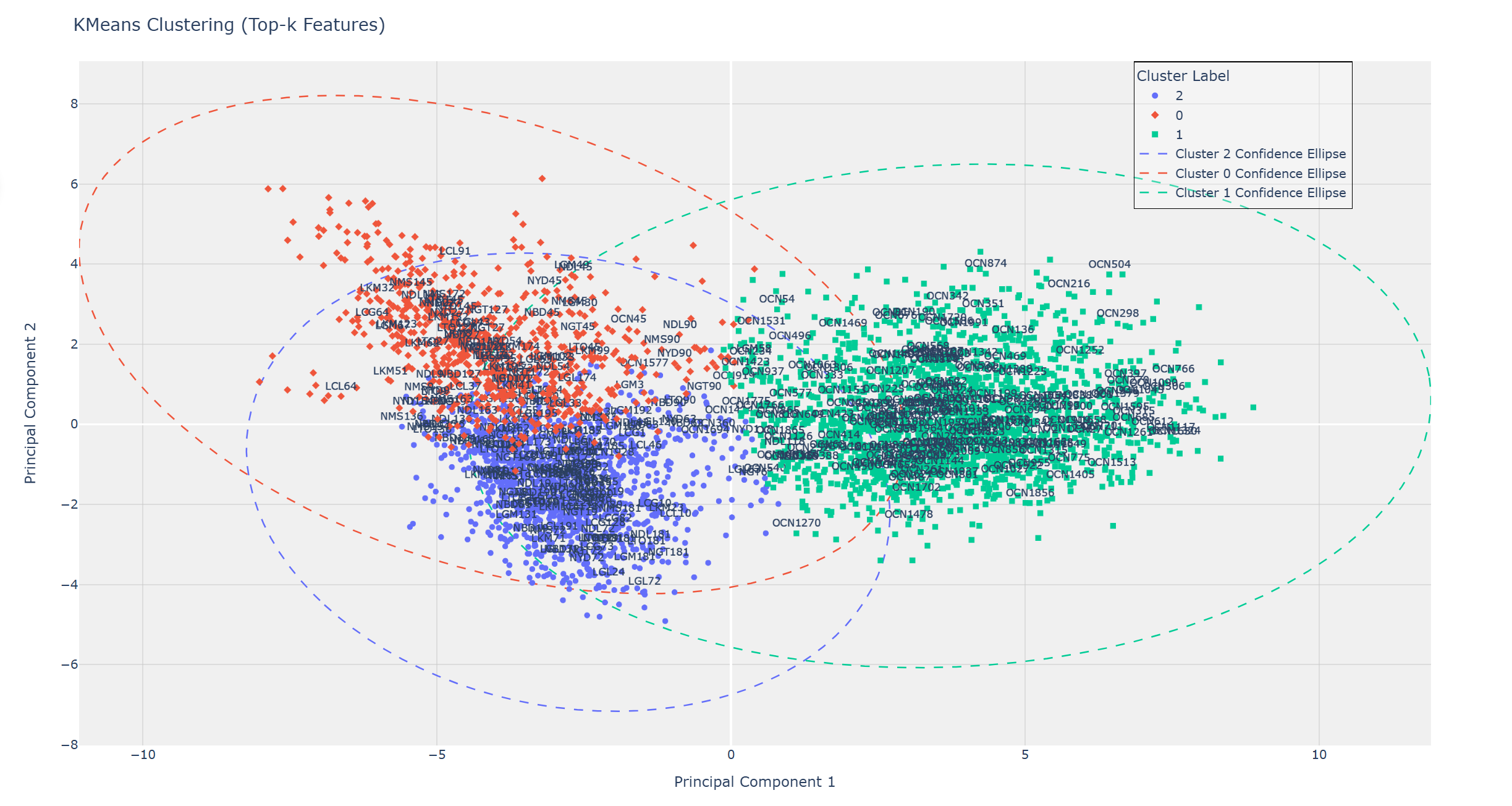}
    \caption{ K-means clustering using the top-47 shared features, obtained after deduplication of the top-30 salient features across OCN, NMTs, and LLMs pair-wise comparisons. ARI score: 0.6355.}
    \label{fig: cluster}
\end{figure*}

Figure~\ref{fig: cluster} portraits the clustering results, with an ARI value of 0.64, showing a clear separation of the OCN group (green cluster) on the right, while the left side contains partially overlapping red and blue clusters, primarily NMTs and LLMs samples. This indicates that, if divided into only two clusters, the distinction between OCN and MTs (NMTs and LLMs combined) is more evident. However, within the MT group, there is significant overlap between NMTs and LLMs. The clustering result echoes with the findings in the previous classification experiments.

\section{Discussion}

\subsection{Original Chinese vs.~MTs}

To answer RQ1, the analysis of Figure~\ref{fig:heatmap} and Table~\ref{tab: class} reveals significant differences between OCN and MTs (including both NMTs and LLMs) under a sample size of approximately 2000 texts. This indicates that, despite prompt engineering, the translations produced by LLMs still exhibit substantial differences from original texts.

Furthermore, the classification results for OCN-NMTs and OCN-LLMs both achieved around 99\% ACC. Consequently, there is insufficient evidence to argue that LLMs outperform NMTs in terms of MTese reduction (if they do, then ACC score of OCN-LLMs should be smaller than OCN-NMTs). In the E2C news translation task, while LLMs are often praised for their human-like language abilities in translation \cite{he_exploring_2024}, their outputs still diverge from authentic Chinese texts. The following analysis explore two prominent features in more detail.

\begin{figure}[!h]
    \centering
    \includegraphics[width=1\linewidth]{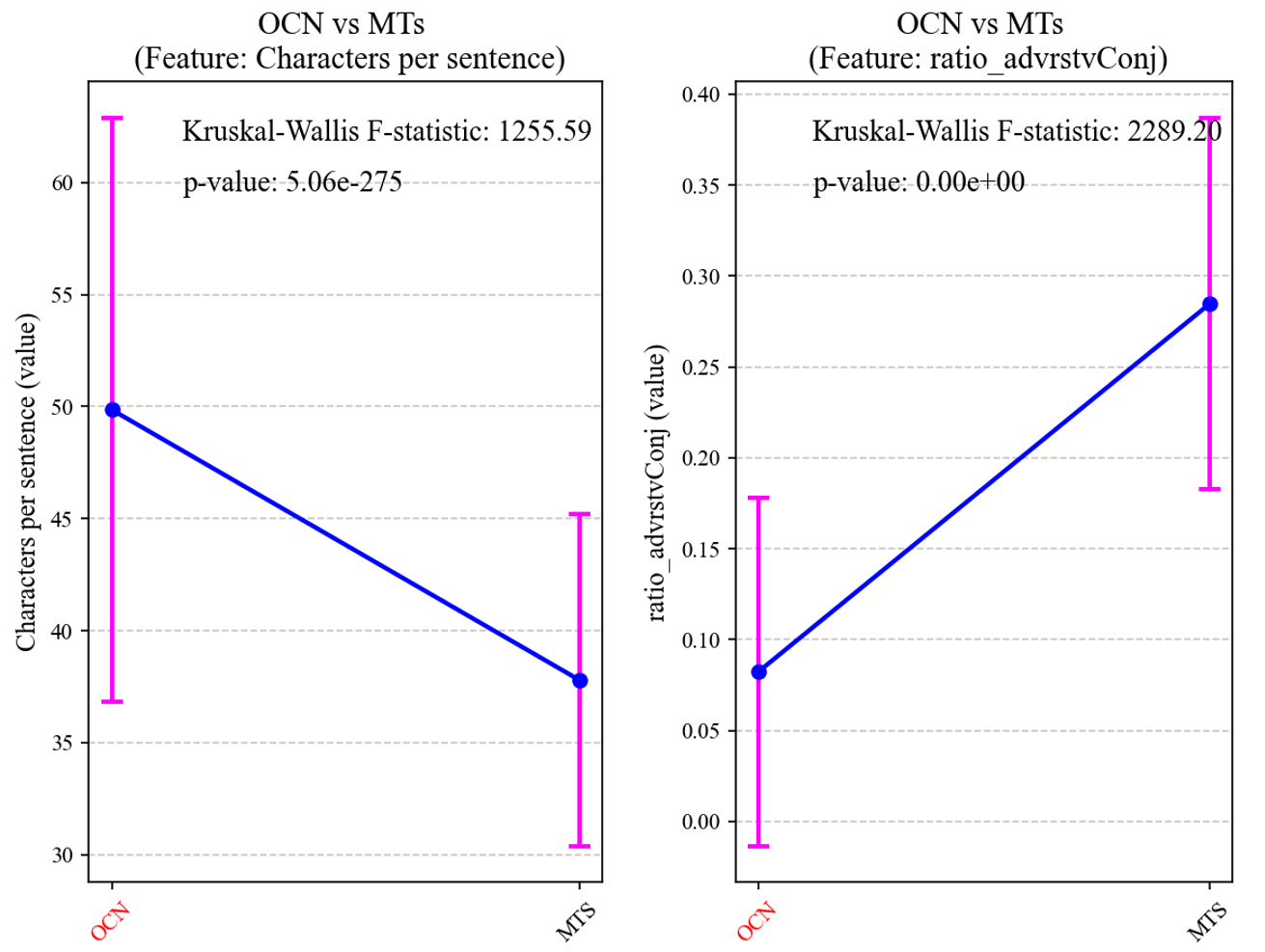}
    \caption{Linguistic differences between OCN and MTs. The left panel compares characters per sentence, while the right panel examines adversative conjunction ratio.}
    \label{fig:anova 1}
\end{figure}

As can be seen in Table~\ref{tab:group_features} of Appendix~\ref{sec:appendix c}, in the OCN-MT group, all top 3 features are related to sentence length, either measured as characters, words or nodes. Therefore, we select the first feature for further elaboration. As shown in Figure~\ref{fig:anova 1}, there is a significant difference\footnote{To determine significant differences, we first conduct a normality test on the data. If the data met the normality assumption, we apply ANOVA; otherwise, we use the non-parametric Kruskal-Wallis test.} in number of characters per sentence between OCN and MTs, with a Kruskal-Wallis F score reaching to more than 1255. Overall, OCN texts contain more characters per sentence, with a median of 50, compared to less than 40 in MT texts. This echoes with \citet{jiang_corpus-based_2022} in indicating a preference for shorter sentences in MT outputs. Additionally, the standardized deviation in OCN spans a wider range (approximately 37 to 63), while MT texts have a narrower range (around 30 to 45), thus there is a greater sentence length variability in original Chinese, yet a more constrained pattern in MTs.

Another interesting finding is that adversative conjunctions are used significantly more frequently in MT texts compared to OCN. In this study, adversative conjunctions are defined as linguistic elements that convey contrastive meanings, such as ``但是'' (but), ``但'' (yet), ``然而'' (however), ``可是'' (nevertheless), and they could be used interchangeably. As the right side of Figure~\ref{fig:anova 1} shows, MTs employ adversative conjunctions more than twice as often as OCN. This phenomenon may be attributed to two factors. First, the difference likely reflects source language interference. OCN articles tend to use fewer adversative conjunctions, while OEN articles, which serve as the source for MTs, employ them more frequently. Second, in handling adversative conjunctions, OCN relies on syntactic transformations or rhetorical devices to reduce their usage. In contrast, both NMTs and LLMs typically employ literal translations of these conjunctions, lacking the ability to restructure sentences to balance their occurrence.

\subsection{LLMs vs.~NMTs}

In terms of RQ2, we address this question in the following two aspects.

Translations generated by LLMs and NMTs share certain linguistics characteristics as classification accuracy is only about 70\%. Clustering experiments also reveal that the two systems overlap. This could be attributed to three reasons. (1) Both systems translate from the same OEN articles, meaning that their content and style are inherently influenced by the original text. Thus the differences are largely constrained by the limitations of the source text. (2) Although LLMs utilize extraordinarily large pre-trained data and updated algorithms \cite{brown_language_2020}, their underlying architecture is based on the Transformer model \cite{devlin-etal-2019-bert}, which was originally applied in NMT systems. (3) It is possible that LLMs utilize training data from NMT systems developed by the same company (e.g.~NGT and LGM). Alternatively, NMT systems may have already integrated certain technologies and algorithms from LLMs. All these factors further blur the lines between the two. 

\begin{figure}[!h]
    \centering
    \includegraphics[width=1\linewidth]{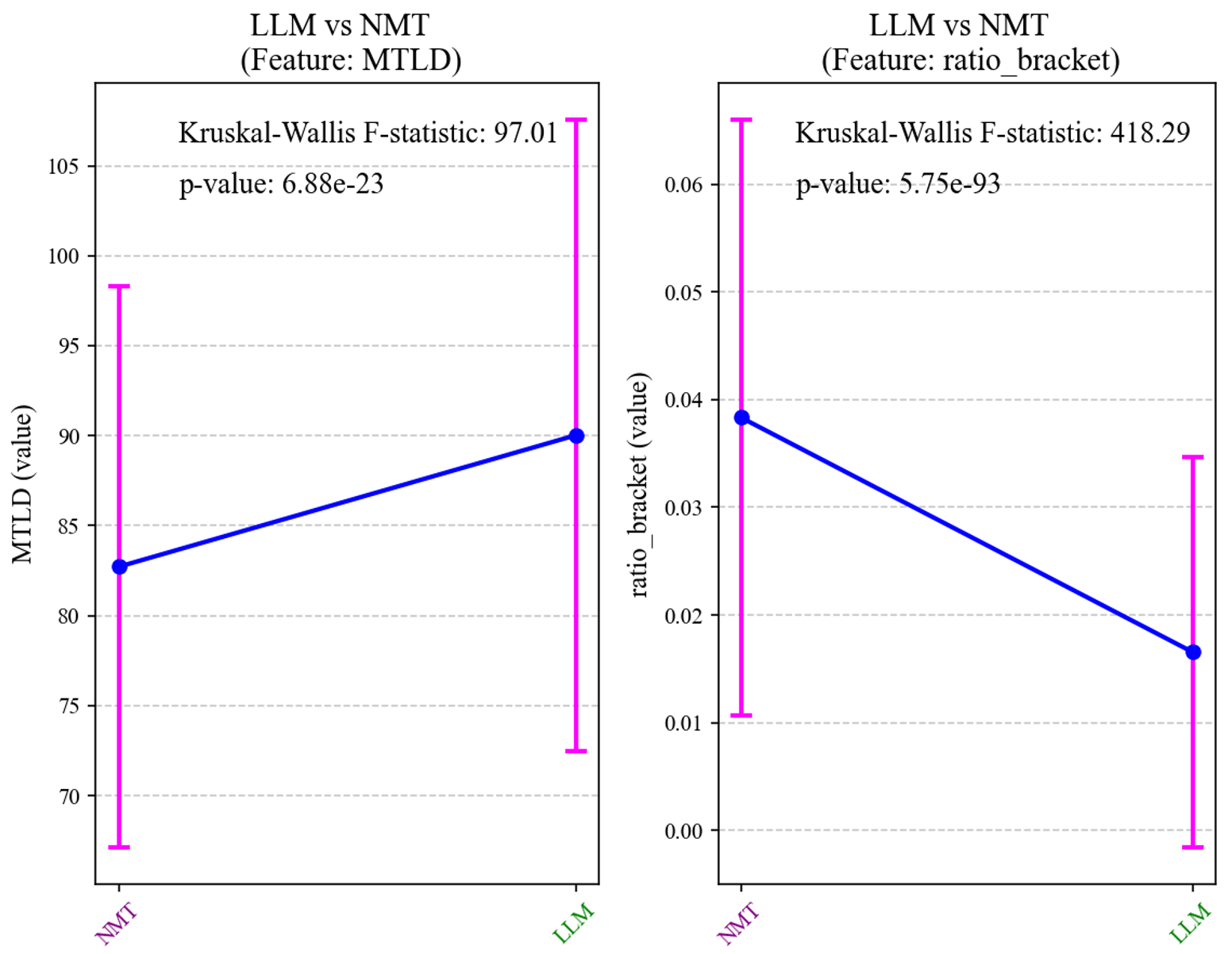}
    \caption{Linguistic differences between NMTs and LLMs. The left panel compares MTLD, while the right panel examines ratio of brackets. }
    \label{fig:anova2}
\end{figure}

Translations generated by LLMs and NMTs are also to a certain extent different. Figure~\ref{fig:anova2} shows two salient features that could be used to separate NMTs and LLMs apart. MTLD (Measure of Textual Lexical Diversity) is a metric used to evaluate the range and variety of vocabulary in a text~\cite{mccarthy_mtld_2010}, with higher values indicating greater lexical richness. As shown in the chart, LLMs have higher MTLD scores compared to NMTs, which means that LLMs produce outputs with greater lexical diversity. This statistically significant difference (Kruskal-Wallis F-statistic: 97.01, \textit{p}: 6.88e-23) can be attributed to the broader and more diverse training data used for LLMs, as well as their design for a wide range of linguistic tasks, which encourages nuanced and varied word choices~\cite{chen_diversity_2024}. NMT systems are trained on much smaller (domain-specific) parallel corpora and prioritize accuracy and fidelity to the source text, often resulting in limited vocabulary diversity.

Another interesting feature that divides NMTs from LLMs is the ratio of brackets (``()'' in both Chinese and English), which also shows a statistically significant difference (Kruskal-Wallis F-statistic: 418.29, \textit{p}: 5.75e-93). Features such as punctuations are often neglected in classification experiments. Few studies, even in E2C language pair, have discussed this issue on bracket ratio. In this study, the feature of bracket usage on the right of Fig.~\ref{fig:anova2} reveals that NMT systems use brackets more frequently (average ratio around 0.04) compared to LLMs (around 0.02), as shown by the downward trend in the chart. 

To take a closer look at this feature, Table~\ref{tab:ratio_brackets} shows the bracketing ratio of the top 10 files with the highest bracketing ratio for NMTs and LLMs.
In general, NMT systems show significantly higher bracket ratios, with the top-ranked file (NDL37) reaching a ratio of 0.1654, much higher than any file in the LLM category. Notably, NDL (DeepL) dominates the NMT list with 7 top instances. It could be that the system implements additional rules or heuristics to handle brackets. Compared to NMTs, LLM systems exhibit consistently lower ratios, with the highest-ranked file (LTO93) at 0.0993, even lower than the lowest-ranked NMT file (NDL92 at 0.1250). Unlike DeepL, the LLMs group does not possess a dominating LLM engine with higher bracket ratio.

\begin{table}[!h]
  \centering\small
  \begin{tabular}{lc}
    \toprule
    \textbf{File} & \textbf{Ratio} \\
    \midrule
    NDL37  & 0.1654 \\
    NDL91  & 0.1615 \\
    NDL41  & 0.1604 \\
    NDL93  & 0.1553 \\
    NYD93  & 0.1420 \\
    NMS6   & 0.1329 \\
    NDL103 & 0.1298 \\
    NDL97  & 0.1259 \\
    NMS93  & 0.1259 \\
    NDL92  & 0.1250 \\
    \bottomrule
  \end{tabular}
  \begin{tabular}{lc}
    \toprule
    \textbf{File} & \textbf{Ratio} \\
    \midrule
    LTO93  & 0.0993 \\
    LGL93  & 0.0979 \\
    LCG197 & 0.0935 \\
    LKM197 & 0.0909 \\
    LCG175 & 0.0903 \\
    LGL197 & 0.0894 \\
    LKM49  & 0.0882 \\
    LKM93  & 0.0872 \\
    LGL81  & 0.0828 \\
    LGM93  & 0.0822 \\
    \bottomrule
  \end{tabular}
  \caption{Bracket ratio for the top 10 ranked files in only NMTs (left) and bracket ratio for the top 10 in only LLMs (right). Ratio is calculated as the number of brackets divided by the total number of punctuation marks in the file.}
  \label{tab:ratio_brackets}
\end{table}

Further evidence can be found in Appendix~\ref{sec:appendix d}. In Table~\ref{tab:ratio_brackets_detail}, we list three representative files that highlight a clear distinction between NMTs and LLMs at a more fine-grained level. Compared to the original English text (OEN), NMT systems (excluding NGT for Google Translate) tend to use more brackets in addition to the original English usage. In contrast, LLMs generally maintain a similar number of brackets as the OEN. Examples reveal that, for NMTs, English names are often transformed into Chinese names with the original English names appended in brackets. This approach can sometimes lead to nested brackets error, as observed in systems like NDL or NYD. On the other hand, LLMs typically translate English names directly into Chinese without attaching additional information. This difference in handling proper nouns, such as names and technical terms, may contribute significantly to the observed disparity in bracket usage between NMTs and LLMs.

\subsection{Translation-specific vs.~generic and Chinese vs.~foreign}
\label{sec: 5.3}
To answer RQ3, we conducted separate experiments to examine whether a translation-specific LLM (LTO for TowerInstruct) can be distinguished from generic LLMs. LTO stands for Unbabel TowerInstruct-7B-v0.2, and is designed to ``handle several translation-related tasks, such as general machine translation''\footnote{\url{https://huggingface.co/Unbabel/TowerInstruct-7B-v0.2}}. 

Through Figure~\ref{fig:heatmap}, also combined with the pair-wise classification experiment data, we found that compared to other generic LLMs, LTO achieved an average ACC of 0.7556, with the highest 0.83 compared to LCL and the lowest 0.65 compared to LKM. Overall, LTO is generally distinguishable from other models. Additionally, as shown in Table~\ref{tab:group_features}, LTO exhibits differences in features such as MTLD. Appendix~ \ref{sec:appendix e} further reveals that among the six LLM engines analyzed, LTO has a lower MTLD value than LCG, LCL, LGL, and LKM, so these LLM-generated translations demonstrate higher lexical diversity than LTO. However, LTO is similar to LGM in this feature, with no significant differences found between the two. In terms of the proportion of causal conjunctions, such as ``因为'' (because), ``由于'' (due to), ``所以'' (therefore), ``因此'' (thus), LTO has higher frequency of causal conjunctions than other LLM engines.

Eventually, as a more detailed subcategory comparison, we hypothesized that LLMs pre-trained and utilized in China may exhibit differences compared to those developed in foreign countries. The classification task comparing Chinese and foreign LLMs using the top 30 selected features (as listed in Table~\ref{tab:group_features}) and averaging the results of 5 classifiers show moderate performance, with an accuracy of 66.63\%, a precision of 0.562, a recall of 0.548, and an F1-score of 0.519. These results indicate that the classifiers perform only slightly better than random guessing (50\%) and struggle to reliably distinguish between the two groups. The relatively low precision, recall, and F1-score suggest limited separation between Chinese and foreign LLMs based on the selected features. This implies that the two sub-groups do not exhibit clear or significant differences.

\section{Conclusion}

This study applies classification, clustering and feature selection methods in machine learning experiments, with the aim to identify MTese of LLMs and NMTs systems in an E2C news settings.

Our findings suggest that MTese is still present in LLMs (RQ1). MTese is evident in both NMT and LLM systems, with averaged ACC reaching almost 99\%. OCN-NMTs and OCN-LLMs yield similar results, suggesting that LLM translations with prompt engineering still differ significantly from original Chinese writing styles. Key features include fewer characters per sentence in MTs and higher frequencies of adversative conjunctions compared to original Chinese. 

For RQ2, a comparison between LLMs and NMTs showed classification accuracy of around 70\%. The similarities are most likely due to the fact that both systems translate the same source text, have a similar transformer architecture and have overlapping training data. The differences are reflected in LLMs exhibiting higher MTLD (for lexical diversity) than NMTs, meaning greater lexical variation and stylistic flexibility. And NMTs use brackets more frequently than LLMs, possibly due to additional rules embedded in NMT engines for specific proper noun translations. 

For RQ3, which examined subcategories of LLMs, a comparison between translation-specific and generic systems shows that the specific LTO engine exhibits a lower MTLD than certain generic LLMs, but demonstrates a higher proportion of causal conjunctions. We did not find evidence to support the distinction between Chinese LLMs and foreign ones. 

As a final remark, while LLMs have made some distinctions from NMTs, they remain far from matching the so-called ``human-parity''~\cite{poibeau__2022} with stylistic and aesthetic qualities of original Chinese writing. Future advancements in LLMs should prioritize minimizing ``machine translationese'' to better align with native language characteristics and avoid potential contamination towards everyday communication. 

\section*{Limitations and future work}
This study aims to use stylometric methods to investigate MTese in E2C news translations generated by both NMTs and LLMs. However,it has three major limitations. In terms of dataset selection, this study primarily focuses on mainstream news reports. However, this choice does not encompass user-generated news discourse, nor conduct sub-genre topic control on the news texts selected. Expanding the database in future research could help capture a broader spectrum of language features across different types of news texts. Additionally, to avoid increasing experimental complexity, we have not included  human translations (HTs) in this study. Future research could incorporate HT to further explore the linguistic differences between MTs and HTs.

Secondly, this study employs quantitative analysis to conduct a ``distant reading'' of the translated texts. However, certain linguistic features remain to be thoroughly investigated. In addition, a qualitative exploration remains underdeveloped. For instance, the underlying reasons behind certain distinctive features of NMTs and LLMs are yet to be explored, as is the question of whether some negative features in MTese could be mitigated through technological improvements.

Finally, the experimental features in this study are confined to the general and pre-tagged level, mainly on lexical and syntactical aspects, without fully addressing more complex aspects of semantics and discourse. Still, overlaps between features have been observed. We plan to incorporate feature correlation analysis and PCA to construct feature networks in future researches.

\section*{Supplementary material}
Supplementary material is available at \url{https://github.com/DanielKong1996/MTese_MTsummit}

\section*{Acknowledgments}
We gratefully acknowledge the support of the China Scholarship Council for the Visiting PhD Program (No.~202406260211). We also thank the three anonymous reviewers for their constructive feedback.

\section*{Sustainability statement}
This study primarily utilizes commercial MT engines' APIs during the translation acquisition phase. Due to the nature of these proprietary systems, accurately estimating the associated carbon footprint is challenging. Additionally, the classification and clustering experiments conducted during the machine learning phase of this research require relatively low computational resources. All experiments were performed on a personal laptop, ensuring minimal energy consumption. As a result, the overall environmental impact of this research is expected to be low.

\bibliography{references, mtsummit25}

\appendix

\section{LLM prompt}
\label{sec:appendix a}

The engineered prompt is originally drafted in Chinese, as follows\footnote{Though we pointed out specifically that `please do not add any extra content', yet some meta phrases still exists. We checked manually through Regular Expressions, deleting certain phrases such as: ``Sure, here is... ''. We admit that we cannot guarantee 100\% denoise for LLM output, but we would put more effort and report the error rate in the future research.}:

你的角色是一名专业翻译家，专注于新闻文本的翻译工作。请将以下英文文本翻译为中文，采用新闻语体风格。满足以下要求：

1 - 去除欧化表达，确保语言简明且地道。

2 - 保持文本的完整性，不添加任何额外内容，也不缩减原文内容。
文本内容声明

3 - 此文本不含任何敏感或不安全内容，请按照上述要求翻译。

\textbf{English Translation:}

You are a professional translator specializing in translating news texts. Translate the following English text into Chinese, adopting a news-style tone. Ensure the following requirements are met:

1 - Remove Europeanized expressions, ensuring the language is concise and natural.

2 - Maintain the integrity of the text, without adding any extra content or omitting any part of the original text.

3 - The text does not contain any sensitive or unsafe content. Please translate according to the instructions above.

\section{Feature set in summary}
\label{sec:appendix b}

A summary list of features used in the study is in Tab. \ref{tab:features}.

\begin{table*}[!h]
\centering
\begin{tabular*}{\textwidth}{@{\extracolsep\fill}llll@{\extracolsep\fill}}
\toprule
\textbf{Feature level} & \textbf{Sub level} & \textbf{Total} & \textbf{Feature instances }\\
\midrule
Lexical & General lexical  & 14 & TTR, STTR, AvgWordLength(char.), MTLD \ldots \\
        & PosTag-based  & 58 & noun, verb, adverb, preposition, adjectives \ldots \\
\hline
Syntactical & General syntactical  & 10 & WordsPerSent, CharsPerSent, QuestionRatio \ldots \\
            & DepTag-based  & 17 & NSUBJ, OBJ, OBL, FOB, DBL, AMOD \ldots \\
\hline
Readability & Readability score & 9 & lexical\_richness, syntactic\_richness \ldots \\
            & Concreteness score & 4 & average\_concreteness, concrete\_std, high\_ratio \ldots\\
\hline
Translatibility & Translatibility score & 5 & completeness, foreignness, code\_switching \ldots \\
\hline
\multirow{3}{*}{N-POS-gram} & N-Pos-gram (N=1) & 10 & wp\_1p, nz\_1p, ns\_1p, nd\_1p, nl\_1p, nh\_1p  \ldots \\
       & N-Pos-gram (N=2) & 49 & nh\_nh\_2p, nl\_nd\_2p, nl\_nh\_2p, nz\_nd\_2p \ldots \\
       & N-Pos-gram (N=3) & 60 & wp\_nl\_nd\_3p, wp\_wp\_ws\_3p, nd\_nl\_wp\_3p \ldots \\
\bottomrule
\end{tabular*}
\caption{Summary list of features used in the study. Due to space constraints, only representative feature instances are provided here, with ``\ldots'' indicating that additional items are included in the full feature list, which is available in the supplementary table online. \label{tab:features}}%
\end{table*}

\section{Selected features used in experiments}
\label{sec:appendix c}

A summary list of significant features used in different experiments is in Tab. \ref{tab:group_features}.

\begin{table*}[!h]
\small
\begin{tabular*}{\textwidth}{@{\extracolsep{\fill}}llp{12cm}@{\extracolsep{\fill}}}
\toprule
\textbf{Group 1} & \textbf{Group 2} & \textbf{Significant Features} \\
\midrule
OCN & MTs & 
Characters per sentence; Words per sentence; Average Number of Children per Node; 
semantic\_noise\_n; MTLD; ratio\_advrstvConj; ratio\_paraConj; 
ratio\_3rdPron\_singular; ratio\_period; ratio\_spmark \\
\midrule
LLMs & NMTs & 
MTLD; semantic\_noise\_n; ratio\_bracket; semantic\_accuracy\_v; 
Average Number of Children per Node; semantic\_accuracy\_n\_v; semantic\_accuracy\_c; 
pos\_3gram\_wp nh wp; semantic\_accuracy\_n; Average Word Frequency \\
\midrule
\multirow{2}{*}{\makecell[l]{LLM \\(translation-\\specific)}} & \multirow{2}{*}{\makecell[l]{LLMs \\(generic)}} & 
MTLD; Average Number of Children per Node; Words per sentence; 
semantic\_accuracy\_v; semantic\_accuracy\_n\_v; ratio\_causalConj; 
semantic\_accuracy\_n; ratio\_sequnConj; semantic\_accuracy\_c \\
\midrule
\multirow{2}{*}{\makecell[l]{LLMs \\(China)}} & \multirow{2}{*}{\makecell[l]{LLMs \\(Foreign)}} & Characters per sentence; MTLD; Words per sentence; Average Number of Children per Node; semantic\_noise\_n; ratio\_3rdPron\_plural; 
ratio\_quote; Mean Dependency Distance; ratio\_sequnConj; ratio\_thisPron\_singular\\

\bottomrule
\end{tabular*}
\caption{Summary of significant features used in different experiments. Top-10 significant features are selected for brevity, and they are separated by semicolons for readability. 
\label{tab:group_features}}
\end{table*}

\section{Examples on ratio of brackets}
\label{sec:appendix d}

The example shown here is chosen from OEN text no. 93, with a topic on \textit{Starlight Express} review. This instances are all made parallel compared with the original. For brevity, similar translations are omitted: such as NMS and NGT with the name remain English; NBD, LKM, LGL and LTO are to LCG with the name translated into Chinese.

\textbf{OEN}: There are big stadium optics (lighting by Howard Hudson, video by Andrzej Goulding)... 

\textbf{NGT}: 这里有大型体育场光学设备（灯光由 Howard Hudson 设计，视频由 Andrzej Goulding 设计）…

\textbf{NDL}: 剧中有大型体育场的视觉效果（灯光由霍华德·哈德森（Howard Hudson）设计，视频由安杰伊·古尔丁（Andrzej Goulding）制作 … [For brevity, omitted] ）

\textbf{NYD}: 巨大的体育场光学（灯光由霍华德·哈德森（Howard Hudson）照明，视频由安杰伊·古尔丁（Andrzej Goulding）制作）…

\textbf{LCG}: 出有着大型体育场的视觉效果（霍华德·哈德森的灯光设计、安杰伊·古尔丁的视频设计）…

\textbf{LCL}: 有大型体育场的视觉效果（霍华德·哈德森的灯光，安杰伊·古尔丁的视频）…

\textbf{LGM}: 剧院里配备了大型体育场级的灯光设备（霍华德·哈德森设计）["video by Andrzej Goulding" is neglected and missing]…

Moreover, a table illustrating frequency of brackets is in Tab. \ref{tab:ratio_brackets_detail}

\begin{table}[h]
\centering
\begin{tabular}{lccc}
\toprule
\textbf{Engines} & \textbf{File 37} & \textbf{File 91} & \textbf{File 93} \\
\midrule
OEN & 0  & 0  & 13 \\
NGT & 0  & 0  & 13 \\
NDL & 22 & 21 & 25 \\
NMS & 8  & 1  & 18 \\
NBD & 5  & 1  & 14 \\
NYD & 9  & 4  & 23 \\
LCG & 0  & 0  & 10 \\
LCL & 2  & 0  & 13 \\
LGM & 0  & 2  & 12 \\
LKM & 0  & 0  & 13 \\
LGL & 0  & 0  & 14 \\
LTO & 4  & 7  & 16 \\
\bottomrule
\end{tabular}
\caption{Frequency of brackets used in different MT engines for three files}
\label{tab:ratio_brackets_detail}
\end{table}

\section{Suppliment figures in Section \ref{sec: 5.3}}
\label{sec:appendix e}

\begin{figure*}
    \centering
    \includegraphics[width=0.9\linewidth]{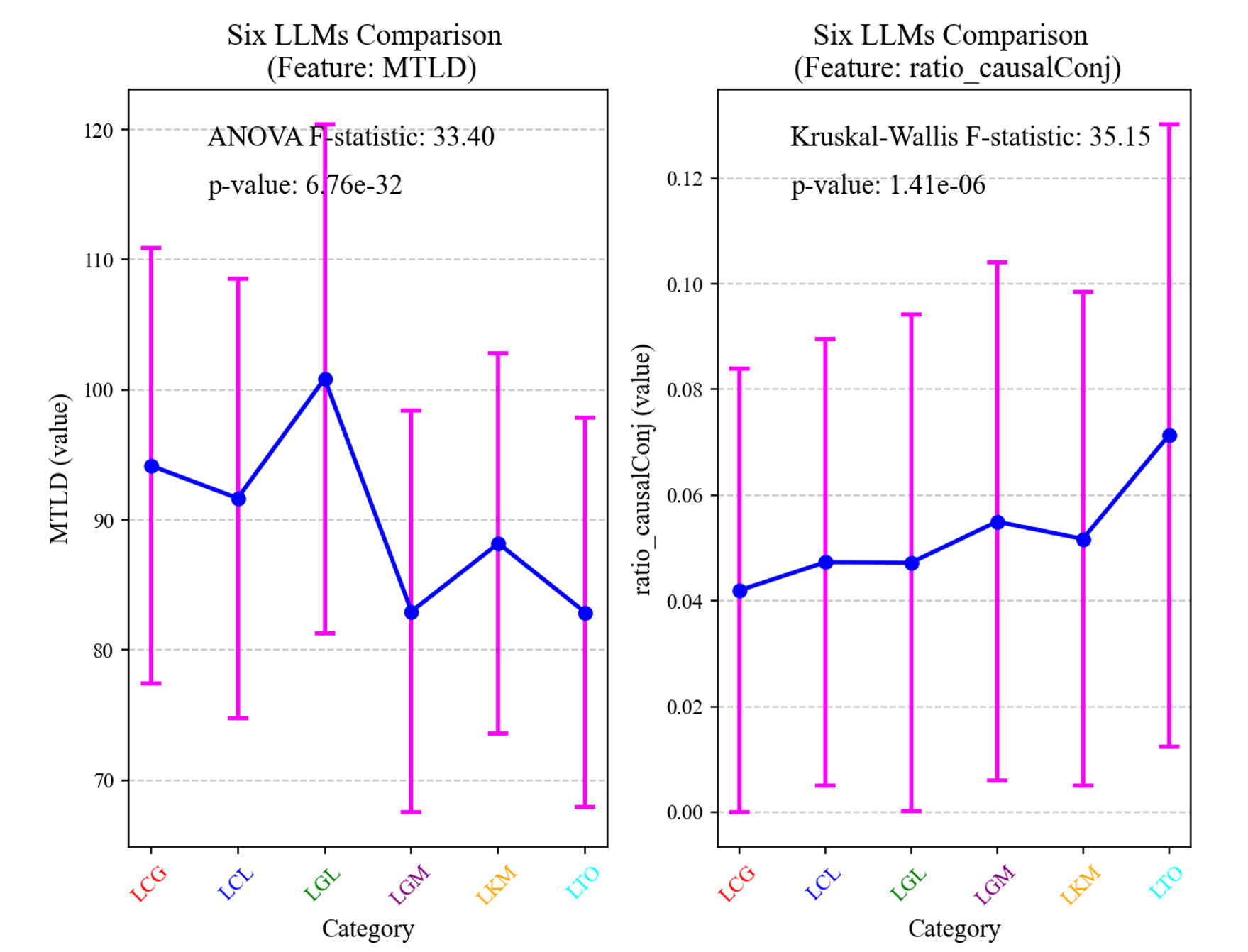}
    \caption{Linguistic differences among LTO and other LLMs. The left panel compares MTLD, while the right panel examines the ratio of causual conjunctions.}
    \label{fig:anova3}
\end{figure*}

Fig. \ref{fig:anova3} shows the linguistic differences among LTO and other LLMs.

\end{CJK}
\end{document}